\pdfoutput=1
\documentclass[11pt]{article}

\usepackage[preprint]{acl}

\usepackage{times}
\usepackage{latexsym}
\usepackage{amsmath}
\usepackage{amssymb}
\usepackage{enumitem}
\usepackage{longtable}

\usepackage[T1]{fontenc}
\usepackage[utf8]{inputenc}

\usepackage{microtype}

\usepackage{inconsolata}

\usepackage{graphicx}

\title{Uniform Information Density and Syntactic Reduction: Revisiting \textit{that}-Mentioning in English Complement Clauses}

\author{Hailin Hao \\
  University of Southern California \\
  \texttt{hailinha@usc.edu} \\\And
   Elsi Kaiser \\
  University of Southern California  \\
  \texttt{emkaiser@usc.edu } \\}

\begin{document}

\maketitle

\begin{abstract}
Speakers often have multiple ways to express the same meaning. The Uniform Information Density (UID) hypothesis suggests that speakers exploit this variability to maintain a consistent rate of information transmission during language production. Building on prior work linking UID to syntactic reduction, we revisit the finding that the optional complementizer \textit{that} in English complement clauses is more likely to be omitted when the clause has low information density (i.e., more predictable). We advance this line of research by analyzing a large-scale, contemporary conversational corpus and using machine learning and neural language models to refine estimates of information density. Our results replicate the established relationship between information density and \textit{that}-mentioning. However, we find that previous measures of information density based on matrix verbs' subcategorization probability capture substantial idiosyncratic lexical variation. By contrast, estimates derived from contextual word embeddings account for additional variance in patterns of complementizer usage.\footnote{Code is available \href{https://github.com/HailinHao/that-Reduction}{here}.}

\end{abstract}

\section{Introduction}

Language production is highly flexible across all levels of linguistic analysis, such as phonetics, lexicon, and syntax. Such flexibility in production enables researchers to ask the question: What cognitive mechanisms guide our choice among competing alternatives? A prominent account, Uniform Information Density (UID; \citealp{jaeger2010redundancy}; \citealp{levy2007informationdensity}), proposes that speakers exploit this flexibility to maintain a consistent rate of information transmission. According to UID, speakers tend to structure their utterances to distribute information as evenly as possible across the linguistic signal to ensure robust information transmission while maintaining efficient use of the communication channel. Following \citeposs{shannon1948mathematical} information theory, the information density of a linguistic unit \( u \) (e.g., a phoneme, a word, or a syntactic structure) given its context is defined as:

\begin{equation}
  \label{eq:information_density}
  I(u) = -\log(P(u \mid \text{context}))
\end{equation}
where \( P(u \mid \text{context}) \) denotes the contextual probability of \( u \). This is also commonly referred to as surprisal (\citealp{hale-2001-probabilistic}; \citealp{levy-2008-expectation}). 

\begin{figure}[!t]
  \centering
  \includegraphics[width=0.9\linewidth]{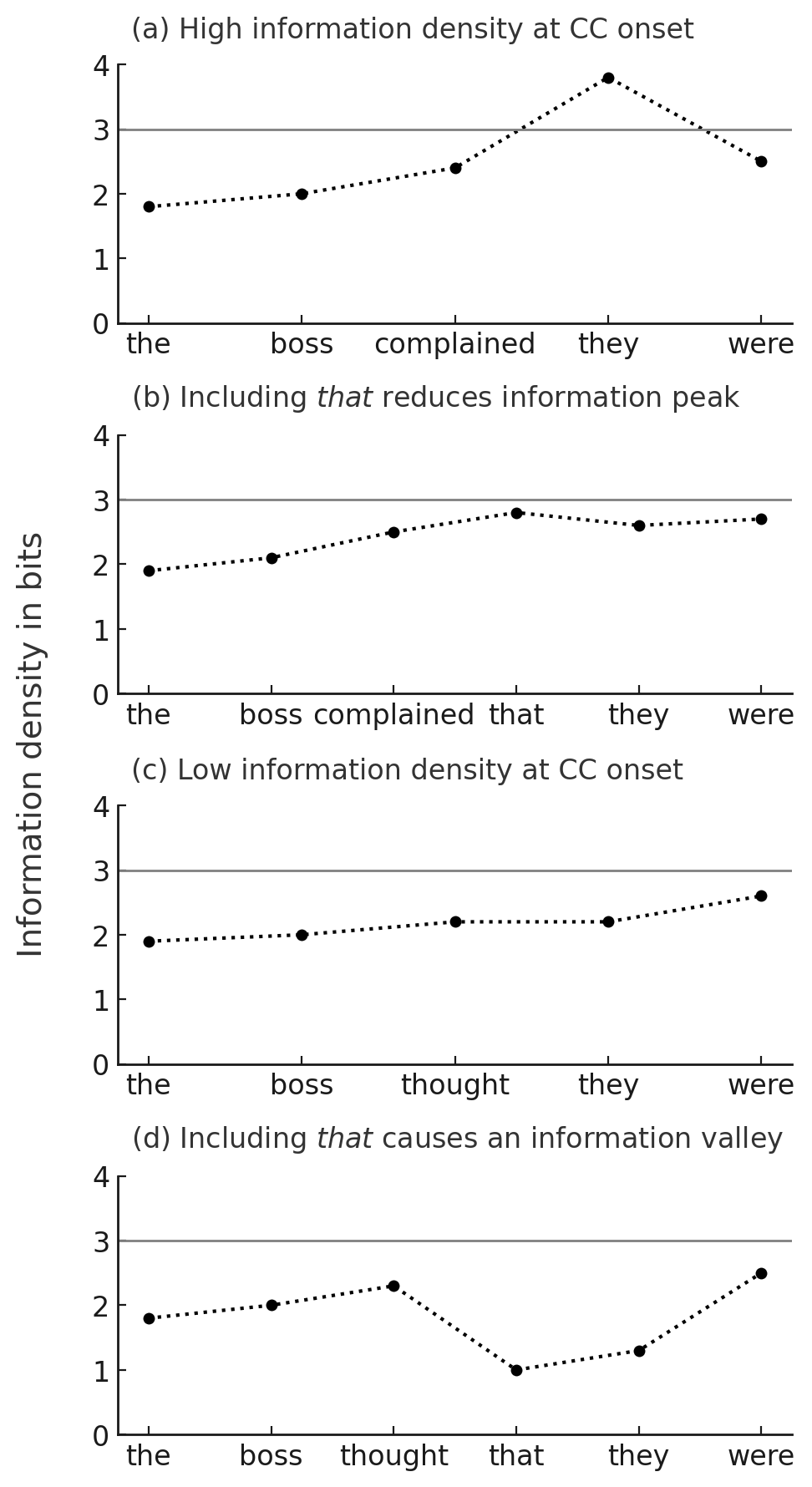}
  \caption{Illustration of per-word information density (note that the solid line represents a hypothetical channel capacity, and that the information density values are schematic only). } 
  \label{fig:uid_illustration}
\end{figure}

In an influential study, \citet{jaeger2010redundancy} demonstrated that UID effects can be observed at the syntactic level. He examined the optional complementizer \textit{that} in English complement clauses (henceafter CCs; e.g., (1)) and found that \textit{that} is more likely to be included when the information density of the CC is high—that is, when the contextual probability of a CC given the preceding context, \( P(CC \mid \text{context}) \), is low. 
\begin{enumerate}[label=(\arabic*)]
  \item The boss complained (that) they were crazy. 
\end{enumerate}
The rationale is that an unpredicted CC would create a spike in information density at the clause onset without \textit{that}, since the CC is unexpected, while including \textit{that} helps smooth the distribution by signaling the upcoming structure. Conversely, when a CC is highly predictable, \textit{that} becomes redundant and may introduce an information density trough. This preference is illustrated in Figure~\ref{fig:uid_illustration}. When the CC onset is information-heavy, potentially exceeding the channel’s capacity, as in Figure~\ref{fig:uid_illustration}a (because CC has low probability after \textit{the boss complained}), including \textit{that} can reduce peak information density (Figure~\ref{fig:uid_illustration}b). In contrast, when the onset is relatively low in information density, omitting \textit{that} results in a smooth information profile (Figure~\ref{fig:uid_illustration}d), while mentioning \textit{that} would create a valley (Figure~\ref{fig:uid_illustration}c).

While \citet{jaeger2010redundancy} provided important evidence for UID, several limitations remain in this work. First, \( P(CC \mid \text{context}) \) was quantified using matrix verbs' subcategorization probabilities—that is, the proportion of times a given verb (based on its lemma) takes a CC as its syntactic object, based on corpus-derived frequencies. This static measure might not not fully capture dynamic predictive processes and may conflate predictability with verb-specific variation. Second, the study used a relatively small and outdated dataset: about 8,000 CCs with 31 matrix verbs from the Switchboard corpus (\citealp{godfrey1992switchboard}; \citealp{marcus1999treebank}), which may limit the generalizability of the findings. Given the theoretical importance of \citeposs{jaeger2010redundancy} findings, a reexamination using larger datasets and more refined predictability measures is needed. 

To address these limitations, in the current work we analyze a modern large-scale corpus called Conversation: A Naturalistic Dataset of Online Recordings (CANDOR; \citealp{reece2023candor}). To preview, we extract over 50,000 unique cases of CCs after data cleaning, encompassing 50 unique matrix verbs. In addition, we incorporate insights from machine learning and neural language models, especially contextual word embeddings, to refine measures of structural predictability. Such refined estimation also allows us to investigate whether improved predictability of CCs leads to better modeling of \textit{that}-mentioning.

\section{Related Work}
\subsection{Psycholinguistic Evidence for UID}
Research supporting the UID hypothesis in language production spans multiple linguistic levels, including phonetics (\citealp{aylett2004smooth}), lexical choice (\citealp{mahowald2013speakers}), syntax (\citealp{jaeger2010redundancy}), and discourse (\citealp{asr2015uid}). For example, past research across many different languages has consistently demonstrated that when a word or phoneme is more predictable in context, it is typically produced with a shorter duration and exhibits reduced phonological and phonetic detail (\citealp{aylett2004smooth}; \citealp{bell2009predictability}; \citealp{cohenpriva2015informativity}; \citealp{pimentel-etal-2021-surprisal}; \citealp{pluymaekers2005lexical}, among others). At the lexical level, \citet{mahowald2013speakers} found that speakers are more likely to use shortened forms of words (e.g., math instead of mathematics) in more predictive contexts. Similarly, at the syntactic level, studies have shown that optional syntactic markers, such as \textit{that} in English CCs (e.g., \textit{I think (that) the weather is very nice}; \citealp{jaeger2010redundancy}) and object relative clauses (e.g., \textit{the groceries (that) they brought home}; \citealp{levy2007informationdensity}), are more frequently omitted when the upcoming syntactic structure is highly predictable. The relationship between information density and syntactic reduction also extends cross-linguistically, such as in subject doubling in French (\citealp{liang2024uid}) and optional indefinite articles in German (\citealp{lemke2017optimal}). 

However, the predictions of UID are not always borne out. For example, \citet{zhan-levy-2018-comparing} found that variation in the use of specific versus general classifiers in Mandarin Chinese is better explained by availability-based production accounts. In addition, \citet{kuperman2007morphological} observed that Dutch interfixes are pronounced longer when they have higher contextual probability, contrary to UID predictions, which they attributed to paradigmatic enhancement. These divergent findings underscore the need for further evaluation of UID.

\subsection{Neural Language Model and Structural Knowledge }
 
A range of studies has probed neural language models' sensitivity to linguistic structures. \citet{linzen-etal-2016-assessing}, for instance, evaluated LSTMs' ability to capture subject-verb agreement using template-based test data. Extending this approach, \citet{warstadt-etal-2020-blimp-benchmark} developed a broader benchmark encompassing a diverse set of linguistic phenomena (see also \citealp{hu-etal-2020-systematic}). Many of these studies rely on surprisal-based evaluations, assuming that ungrammatical continuations should elicit higher surprisal than grammatical ones (e.g., \citealp{futrell-etal-2019-neural}; \citealp{wilcox-etal-2018-rnn}). Other work has adapted stimuli from psycholinguistic experiments, comparing language model surprisal to human behavioral or neural data (\citealp{arehalli2020neural}; \citealp{hao2023transformers}; \citealp{huang2024benchmark}; \citealp{michaelov-bergen-2020-well}). For critical overviews of this literature, see \citet{limisiewicz2020syntax} and \citet{linzen2021syntactic}

Beyond surprisal-based evaluations, researchers have also assessed models' syntactic knowledge through attention head analyses (e.g., \citealp{clark-etal-2019-bert}; \citealp{ryu-lewis-2021-accounting}), meta-linguistic prompting (e.g., \citealp{dentella2024systematic}; \citealp{katzir2023reply}; \citealp{zhou2023syntax}; though see \citealp{hu-levy-2023-prompting}, for critiques of this method), and examinations of contextual word embeddings (e.g., \citealp{li-etal-2022-neural}; \citealp{peters-etal-2018-dissecting}; \citealp{petty2022positionrole}; \citealp{tenney-etal-2019-bert}; \citealp{wilson-etal-2023-abstract}). For instance, \citet{peters-etal-2018-dissecting} demonstrated that contextual embeddings encode a wide range of syntactic information, such as part-of-speech and syntactic boundaries, while \citet{li-etal-2022-neural} showed that contextual word embeddings are sensitive to argument structure even in semantically anomalous sentences.

\section{Structural Predictability Model}

In this section, we detail how we estimate $P(\text{CC} \mid \text{context})$, the structural predictability of CCs. Recall that the information density of a CC is defined as the negative logarithm of $P(\text{CC} \mid \text{context})$. To obtain these estimates, we trained several neural binary classifiers using either hand-selected linguistic features from the pre-CC context or contextual word embeddings of the matrix verb. Hand-selected features offer interpretability and theoretical grounding but may overlook subtle or high-dimensional patterns in the linguistic context. In contrast, contextual word embeddings (e.g., from BERT or GPT models) encode nuanced semantic and syntactic information by capturing how a word’s meaning dynamically changes with its context, but at the cost of interpretability (\citealp{kennedy2021text}). To balance this trade-off, we evaluated models trained on each feature type separately.

\subsection{Linguistic Features} 
We included features from the matrix verb and the matrix subject in the pre-CC context. For the matrix verb, we included its subcategorization probability, estimated from the CANDOR corpus (see Appendix~\ref{sec:verbs}), as well as its log frequency (SUBTLEX; \citealp{brysbaert2009moving}), factivity (i.e., whether it presupposes the truth of the clause it introduces; \citealp{karttunen1971implicative}), tense (base form vs. inflected), and position within the sentence. We also included two features related to the matrix subject: form (\textit{I}, \textit{You}, \textit{Other pronouns} vs. \textit{Other nouns}) and log frequency. 

To identify the most effective feature set, we performed incremental feature selection, adding features one at a time starting from the matrix verb’s subcategorization probability. A feature was retained only if it improved model fit according to Akaike Information Criteria (AIC; \citealp{akaike1974aic}) and Bayesian Information Criteria (BIC; \citealp{schwarz1978bic}), both of which balance model fit and complexity by penalizing the inclusion of unnecessary parameters. We also experimented with Lasso regression (\citealp{tibshirani1996lasso}), where we first fitted a linear regression model using all features simultaneously with an L1 penalty to encourage sparsity in the feature set. Features with nonzero coefficients were then used to predict CC presence. 

\subsection{Contextual Word Embeddings}  
To capture richer predictive cues, we extracted contextual embeddings of the matrix verb from GPT-2 Small (GPT-2 henceforth; \citealp{radford2019gpt2}). Note that this context only includes pre-CC information, not information after the CC onset (e.g., we extracted the embeddings of \textit{complained} from \textit{the boss complained}). GPT-2’s autoregressive architecture enables embeddings based solely on preceding context, aligning with incremental sentence processing. We used the final hidden state of the verb token and reduced the 768-dimensional embeddings to 50 dimensions via PCA (\citealp{jolliffe2002pca}), preserving over 99\% of the variance.

\subsection{Training Data}  
The training data come from the CANDOR corpus (\citealp{reece2023candor}), a large-scale dataset of 1,656 dyadic conversations recorded over Zoom. The corpus is publicly available and can be requested \href{https://betterup-data-requests.herokuapp.com/}{here}. These conversations capture spontaneous, unscripted exchanges between strangers and are supplemented with detailed survey data. The corpus includes 1,456 unique participants representing a diverse range of gender identities, educational backgrounds, ethnicities, and generations. The mean conversation duration is 31.3 minutes (SD = 7.96, min = 20). All analyses in this study are based on existing transcripts from the corpus, totaling approximately 8 million words. Transcripts were segmented using the Cliffhanger algorithm, which groups utterances based on terminal punctuation (e.g., periods, exclamations, questions) and integrates backchannels into broader conversational units.

Transcripts were automatically parsed using spaCy’s dependency parser (\citealp{spacy2}), following Universal Dependencies conventions (\citealp{demarneffe2021universal}; \citealp{nivre2016universal}). We began with 86 matrix verbs that can take CCs, identified by \citet{jaeger2010redundancy} and \citet{jaeger2013information}. Based on frequency in the CANDOR corpus, we selected the 50 most frequent verbs ($\geq 100$ occurrences; see Appendix~\ref{sec:verbs}).

We then extracted all instances of these 50 verbs, regardless of whether they were followed by a CC, direct object, or other dependents. We excluded cases where the verb was sentence-final or the matrix subject was missing. Each instance is labeled as \textbf{1} if followed by a CC and \textbf{0} otherwise. The final dataset consists of 236,504 training examples, with 33.01\% labeled as \textbf{1}. 

\subsection{Model Architecture, Training, and Evaluation} 
We trained feedforward neural networks to predict CC presence. The input features were fed into three hidden layers (128, 64, and 32 units, respectively) with ReLU activation, batch normalization, and 0.2 dropout. The final layer used sigmoid activation to produce probabilities ranging from 0 to 1. Before training, all numerical predictors were z-scored, and categorical variables factor-encoded.

The model was trained using binary cross-entropy loss and optimized with Adam (\citealp{kingma2015adam}) in minibatches of 1024 instances (learning rate = 0.001, weight decay = 1e-5). Training proceeded for up to 50 epochs, with early stopping if validation loss did not improve after five epochs. We used five-fold stratified cross-validation to maintain class distribution across splits. 

\subsection{Structural Predictability Model Results}

Results from the incremental selection of linguistic features are presented in Table~\ref{tab:model_comparison}. Recall that a new feature was added only if it improved model performance in terms of AIC and BIC. Table~\ref{tab:model_comparison} reports the change in AIC and BIC relative to the previously selected model. For reference, we also report each model's F1 score and log loss. As shown in Table~\ref{tab:model_comparison}, including subcategorization probability leads to reductions in both AIC and BIC relative to the baseline model, as well as lower log loss and higher F1 scores. However, none of the additional linguistic features resulted in further improvements according to both AIC and BIC. In fact, the more complex models even show slight decreases in F1 scores. Thus, among the linguistic features considered, only subcategorization probability enhanced the predictions of CC presence.

\begin{table*}[!t]
\centering
\begin{tabular}{lrrrr}
\hline
\textbf{Features} & \textbf{AIC $\Delta$} & \textbf{BIC $\Delta$} & \textbf{F1} & \textbf{Log Loss} \\
\hline
Intercept only & -- & -- & 0.000 & 0.6346 \\
Subcategorization Probability & -13629.10 & -13629.10 & 0.660 & 0.491 \\
+ Verb Frequency & 128.94 & 1456.78 & 0.660 & 0.489 \\
+ Factivity & 320.06 & 1647.90 & 0.660 & 0.491 \\
+ Tense & 156.09 & 1483.93 & 0.654 & 0.490 \\
+ Position & 247.66 & 1575.50 & 0.660 & 0.490 \\
+ Subject Form & -313.63 & 1014.20 & 0.647 & 0.485 \\
+ Subject Frequency & -514.77 & 813.07 & 0.645 & 0.482 \\
\hline
\end{tabular}
\caption{Model comparisons predicting CC presence. Lower AIC, BIC, and log loss, and higher F1 scores indicate better performance.}
\label{tab:model_comparison}
\end{table*}

After applying Lasso Regression, four of seven features were retained: subcategorization probability, verb frequency, factivity, and subject form. Using this refined set, we trained a structural predictability model with the same neural network architecture. However, although this model achieved a lower log loss (0.479), the model showed a decrease in F1 score (0.652) and an increase in BIC compared to the model using only Subcategorization Probability. This result is consistent with earlier findings from incremental feature selection, further confirming that additional linguistic features do not improve predictive performance.

In contrast, the model trained on contextual word embedding features achieved a log loss of 0.442 and an F1 score of 0.691, outperforming all models based on hand-selected features.\footnote{One reviewer suggested experimenting with numerical representations of word meaning that are not sensitive to context, in order to disentangle the contribution of richer vector representations from that of contextual information. To address this, we tested GloVe embeddings \citep{pennington-etal-2014-glove} and obtained an F1 score of 0.66 with a log loss of 0.48. These results did not outperform the subcategorization-only model, suggesting that it is indeed contextual information that drives the improvement.}

Based on these results, we proceed to test how well information density derived from (i) verb subcategorization probabilities and (ii) from contextual word embeddings predicts \textit{that}-mentioning.

\section{Information Density and \textit{that}-Mentioning}

This section reports our statistical models predicting \textit{that}-mentioning. We examine whether higher information density leads to increased \textit{that}-mentioning, as predicted by UID. Recall that in Equation~\eqref{eq:information_density}, CC information density is the negative logarithm of CC structural predictability. Based on the results of the structural predictability models, we experiment with information density derived from verb subcategorization probabilities and contextual word embeddings. Additionally, we assess whether more accurate estimates of CC structural predictability improve the overall fit of models predicting \textit{that}-mentioning.

\subsection{Data} 

As in previous analyses, we relied on parsed transcripts from the CANDOR corpus (\citealp{reece2023candor}). We extracted CCs introduced by the same 50 matrix verbs used for training CC structural predictability models (Appendix~\ref{sec:verbs}) and retained only instances where the matrix verb preceded the CC.

The dataset was further refined based on the following criteria. First, we excluded the first CCs in all conversations (1,656 cases), as we are interested in the potential effects of whether the previous CC is reduced or not. Second, we removed cases lacking either a matrix subject or an embedded nominal subject, as the identity of both the matrix and the embedded subjects are crucial for our analysis (13,076 cases). Lastly, for matrix verbs introducing multiple CCs, only the first occurrence was retained to avoid redundancy (8,097 cases excluded). After exclusions, we are left with 51,276 instances of CCs for analysis.

\subsection{Control variables} 
To rigorously test UID predictions, we controlled for a range of variables that can also affect \textit{that}-mentioning, largely following \citet{jaeger2010redundancy}. We discuss each of them in the following subsections. Importantly, the UID account is not mutually exclusive with these mechanisms. See Appendix~\ref{sec:variables} for a summary of the control variables, including their types, levels, and relative proportions.

\subsubsection{Availability-Based Production} 
According to availability-based accounts (\citealp{bock1985conceptual}; \citealp{ferreira1996syntactic}; \citealp{ferreira2000ambiguity}), optional elements facilitate production when upcoming material is less accessible (i.e., when upcoming material has low frequency). To capture such effects, we included the log frequency of the CC subject head (\textsc{CC Subject Frequency}), the form of the CC subject (\textsc{CC Subject Form}; \textit{I} vs. \textit{You} vs. \textit{Other pronouns} vs. \textit{Other nouns}), and the matrix verb's log frequency (\textsc{Matrix Verb Frequency}). We also included \textsc{Co-referentiality}, a binary predictor indicating whether the matrix and CC subjects are identical (e.g., \textit{I think I...}). 

\subsubsection{Syntactic Priming} Speakers tend to repeat recently encountered structures (\citealp{bock1986syntactic}; \citealp{gries2005syntactic}; \citealp{mahowald2016meta}). We included \textsc{Previous that}, a binary predictor indicating whether \textit{that} was present in the speaker’s or interlocutor’s most recent CC.

\subsubsection{Dependency Locality} Longer dependencies increase production difficulty (\citealp{hawkins2004efficiency}; \citealp{roland2006why}). Three locality measures were considered: \textsc{Matrix Verb-CC Distance} (local vs. non-local), \textsc{CC Subject Length} (number of the CC subject's dependents), and \textsc{CC Remainder Length} (number of words following the CC subject head in the same CC).

\subsubsection{Speaker Commitment} It has been argued that variation in \textit{that}-mentioning is not meaning-equivalent (\citealp{thompson1991grammaticalization}), as sometimes the matrix verb conveys the speaker’s level of commitment rather than introducing a true CC, making \textit{that} unnecessary. Following \citet{jaeger2010redundancy}, we assumed that commitment is highest with first-person subjects, followed by second-person, and then third-person references, and included \textsc{Matrix Subject Form} as a four-level predictor (\textit{I} vs. \textit{You} vs. \textit{Other pronouns} vs. \textit{Other nouns}). 

\subsubsection{Position}  Production difficulty may vary depending on when the CC occur in a sentence. We included \textsc{Verb ID}, the ordinal position of the matrix verb, as a continuous predictor.

\subsubsection{Similarity Avoidance} Speakers may omit \textit{that} to avoid adjacent similar forms if the CC also begins with \textit{that} (\citealp{walter2008constraints}). We included \textsc{that-Doubling} as a binary predictor.

\subsubsection{Disfluencies} Disfluencies can impact syntactic choices (e.g., \citealp{liang2024uid}). We included \textsc{Filled Word} (presence of a filled pause before the CC) and \textsc{Repetition} (immediate repetition of a word, excluding adjectives and adverbs used for emphasis). 

\subsection{Statistical Modeling of \textit{that}-mentioning} 
 Before modeling, all continuous predictors (see Appendix~\ref{sec:variables}) were standardized using z-score normalization. Binary predictors were contrast-coded, and the four-level categorical variables (\textsc{CC Subject Form} and \textsc{Matrix Subject Form}) were coded using successive difference coding: comparing \textit{I} vs. \textit{You}, \textit{You} vs. \textit{Other Pronouns}, and \textit{Other Pronouns} vs. \textit{Other Nouns}.

We fitted a generalized linear mixed-effects model (GLMM; \citealp{jaeger2008categorical}) using the glmer() function from the lme4 package in R (\citealp{bates2015fitting}; \citealp{Rcore2023}), with the presence of \textit{that} as the binary dependent variable. Fixed effects included CC information density and a set of control variables. To account for variability across individuals, we included a random intercept for speaker. In follow-up analyses, we also included a random intercept for matrix verb lemmas (verbs henceforce) to capture verb-specific tendencies in complementizer usage. While these random effects are not directly motivated by theoretical accounts, they serve to control for idiosyncratic variation in baseline rates of \textit{that}-mentioning across speakers and lexical items. Model comparisons were evaluated via AIC and BIC. 

\subsection{Results of \textit{that}-Mentioning}

Here we report the effects of information density on \textit{that}-mentioning to test predictions from the UID hypothesis, alongside other control variables. Information density was estimated using two approaches: the matrix verb’s subcategorization probability and its contextual word embedding. We further examine whether embedding-based estimates—shown to more accurately predict CC presence—better account for \textit{that}-mentioning patterns than verb-based estimates.

\subsubsection{Verb-based Information Density}

The relationship between verb-based information density and \textit{that}-mentioning is illustrated in Figure~\ref{fig:verb_infor_density}. A plot with verb identity can be found in Appendix~\ref{sec:byverb}). As can be seen in Figure~\ref{fig:verb_infor_density}, higher information density is indeed associated with increased rates of \textit{that}-mentioning, consistent with UID predictions, although substantial variability across verbs remains. Results from the statistical model with a speaker random intercept are presented in Table~\ref{tab:regression_alternative}. Generalized Variance Inflation Factors (GVIFs) for fixed effects were close to 1, indicating minimal multicollinearity. Mode results revealed that higher verb-based information density significantly increases the likelihood of \textit{that}-mentioning.

\begin{figure}[t]
  \centering
  \includegraphics[width=0.9\linewidth]{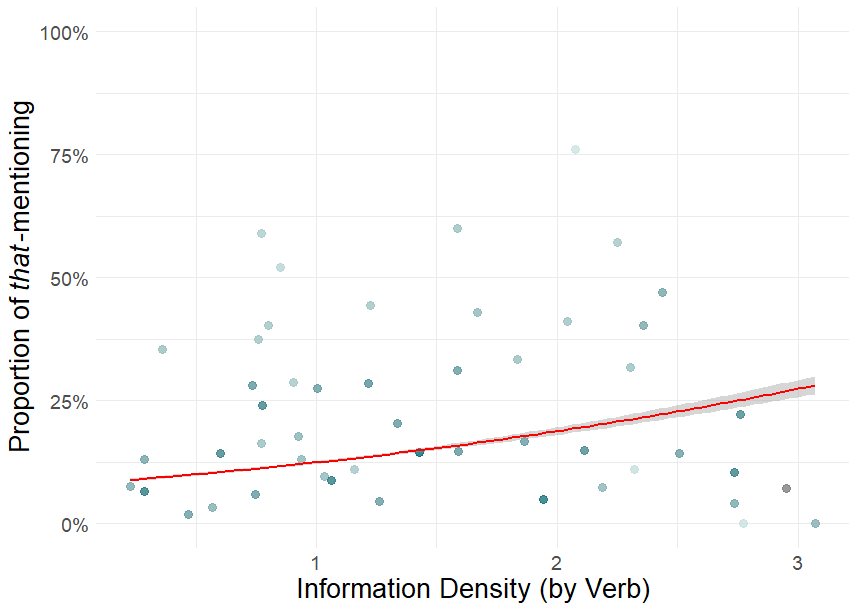}
  \caption{Effects of information density (by verb Subcategorization Probability) on \textit{that}-mentioning. The red line is a logistic regression fit estimated as a binomial GLMM. Each dot represents a verb, while the shading reflects its frequency, with darker shades corresponding to more frequent verbs.}  
  \label{fig:verb_infor_density}
\end{figure}     

Effects of control variables also aligned with several theoretical accounts. First, higher \textsc{CC Subject Frequency} and \textsc{Matrix Verb Frequency} predicted reduced \textit{that}-mentioning, consistent with availability-based accounts. However, \textsc{CC Subject Form} and \textsc{co-referentiality} were non-significant. We also found syntactic priming effects, whereby \textsc{Previous \textit{that}} significantly increased \textit{that}-mentioning. Findings for dependency locality were mixed: longer \textsc{CC Reminder Length} increased \textit{that}-mentioning as expected, but shorter \textsc{CC Subject Length} and \textsc{Matrix Verb-CC Distance} also led to higher \textit{that}-use, contrary to the predictions. Speaker commitement effects were robust—\textit{that} was more likely when the matrix subject was \textit{You} than \textit{I}, with similar trends across other subject types, suggesting \textit{that} signals degrees of speaker commitment. \textsc{Verb ID} had a positive but non-significant effect. Supporting similarity avoidance, potential \textsc{\textit{that}-doubling} reduced \textit{that}-mentioning. Finally, disfluencies measures such as \textsc{Filled word} and \textsc{Repetition} increased \textit{that}-use, with \textsc{Repetition} reaching significance.

\begin{table}[t]
\centering
\begin{tabular}{lrr}
\hline
\textbf{Predictor} & \textbf{Estimate} & \textbf{p-value} \\
\hline
Information Density & 0.28 & $< 0.001$ \\
CC Subject Frequency & -0.16 & $< 0.001$ \\
CC Subject Form 2--1 & -0.00 & $= 0.99$ \\
CC Subject Form 3--2 & -0.02 & $= 0.72$ \\
CC Subject Form 4--3 & 0.03 & $= 0.68$ \\
Matrix Verb Frequency & -0.24 & $< 0.001$ \\
Co-referentiality & -0.05 & $= 0.24$ \\
Previous \textit{that} & 0.23 & $< 0.001$ \\
Matrix Verb-CC Distance & -0.40 & $< 0.001$ \\
CC Subject Length & -0.04 & $< 0.05$ \\
CC Reminder Length & 0.18 & $< 0.001$ \\
Matrix Subject Form 2--1 & 0.69 & $< 0.001$ \\
Matrix Subject Form 3--2 & 0.70 & $< 0.001$ \\
Matrix Subject Form 4--3 & 0.53 & $< 0.001$ \\
Verb ID & 0.02 & $= 0.20$ \\
\textit{that}-Doubling & -0.53 & $< 0.001$ \\
Filled Word & 0.04 & $= 0.36$ \\
Repetition & 0.14 & $< 0.05$ \\
\hline
\end{tabular}
\caption{Regression estimates from the model predicting complementizer presence.}
\label{tab:regression_alternative}
\end{table}

\subsubsection{Embedding-based Information Density}

\begin{figure}[t]
  \centering
  \includegraphics[width=0.9\linewidth]{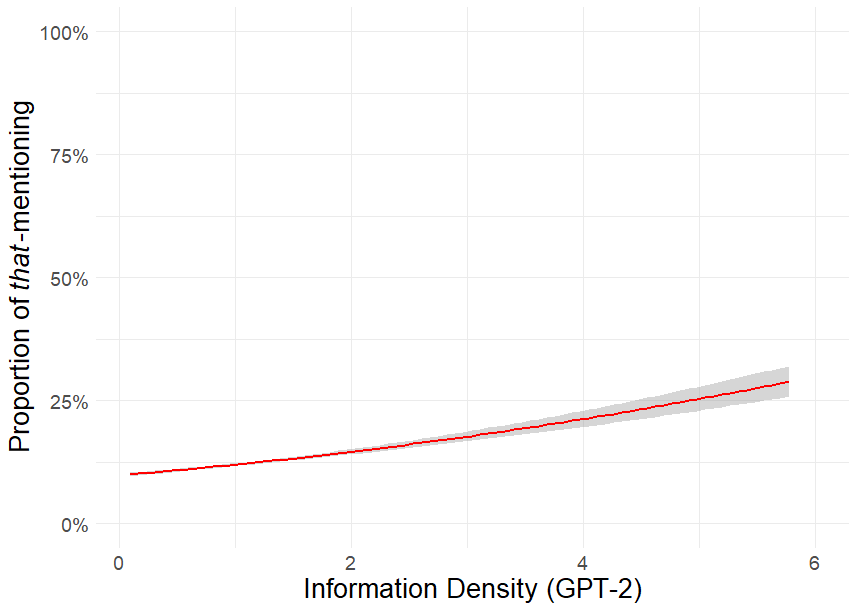}
  \caption{Effects of information density (by contextual word embeddings) on \textit{that}-mentioning. The red line is a logistic regression fit estimated as a binomial GLMM.}  
  \label{fig:embedding_infor_density}
\end{figure}

As shown in Figure~\ref{fig:embedding_infor_density}, embedding-based information density again positively predicted \textit{that}-mentioning. Because the statistical results closely mirrored those of the previous model, we do not report them in detail. Crucially, information density remained a strong predictor ($\beta$ = 0.15; \textit{p} < 0.001).

However, the current model performed worse, with AIC and BIC increasing by 392 and 391 points, respectively, compared to the previous model with verb-based information density. While word embedding features yielded better performance in the structural predictability task, they offered no clear advantage in predicting \textit{that}-mentioning over subcategorization probabilities.

\begin{table*}[t]
\centering
\begin{tabular}{lrr}
\hline
\textbf{Model} & \textbf{AIC} & \textbf{BIC} \\
\hline
Verb-based information density, without verb random intercept & 33344 & 33521 \\
Embedding-based information density, without verb random intercept & 33736 & 33912 \\
Verb-based information density, with verb random intercept & 32302 & 32488 \\
Embedding-based information density, with verb random intercept & \textbf{32277} & \textbf{32462} \\
\hline
\end{tabular}
\caption{Model comparison based on AIC and BIC. }
\label{tab:model_comparison2}
\end{table*}

\subsection{Follow-up Analysis: Verb Random Intercept }
Although the verb-based model initially outperformed the embedding-based model, we were cautious in interpreting this as evidence that embedding-based information density is less effective. In the verb-based model, information density is constant for each matrix verb, potentially conflating information density with verb-specific effects—a limitation of subcategorization probability we mentioned earlier. To address this, we refitted both models with an added random intercept for matrix verbs.\footnote{We note that verb-level predictors remain identifiable even with verb-specific random intercepts, since partial pooling in mixed-effects models prevents perfect collinearity with group-level fixed effects (\citealp{gelman2007data}). In our models, verb-based information density showed no collinearity issues, and verb frequency—a verb-level predictor—remained significant despite the inclusion of verb random intercepts.}

We found that adding a matrix verb random intercept substantially reduced AIC and BIC for both the verb-based and embedding-based models (Table~\ref{tab:model_comparison2}), indicating that a substantial portion of variation in complementizer usage is attributable to verb-specific preferences—patterns tied to individual matrix verbs that were not captured by fixed effects in the previous models.

Additionally, the effects of information density diverged. In the verb-based model, the effect of information density became non-significant ($\beta$ = 0.14; \textit{p} = 0.18), suggesting that its earlier effect was largely driven by verb-specific variation. In contrast, information density estimated from contextual word embeddings remained a significant predictor even after controlling for verb identity ($\beta$ = 0.12; \textit{p} < 0.001). Furthermore, between the two models with verb random intercepts, the embedding-based model showed better fit, reducing AIC and BIC by 25 and 26 points, respectively, suggesting that embedding-based information density captures additional variance in patterns of \textit{that}-mentioning.

\section{Discussion}
This study revisits \citet{jaeger2010redundancy} using a large and modern dataset from the CANDOR corpus. We analyze over 50,000 instances of CCs to test how information density—estimated from different sources—predicts \textit{that}-mentioning, alongside other predictors motivated by alternative theories. We also evaluate whether improved estimates of information density lead to better model performance.

Our results replicate the core finding that higher information density increases the likelihood of overt \textit{that}, as predicted by UID. Information density estimated from verb subcategorization probabilities provides strong predictive power but likely reflected verb-specific preferences rather than a general effect of information density. This is confirmed by follow-up models with random intercepts for matrix verbs, which eliminated the effect of verb-based information density. In contrast, embedding-based information density remains significant in predicting \textit{that}-mentioning, suggesting it captures more abstract, verb-independent information. Moreover, this is consistent with the results of structural predictability models, where GPT-2 embeddings outperform all other features, including verb subcategorization probability, in predicting CC presence, suggesting that it offers a better measure of information density.

 However, we do note that after including the verb random intercept, \citet{jaeger2010redundancy} still find significant effects of verb-based measures of information content. This is likely due to our larger dataset size. Since results of \citet{jaeger2010redundancy} reply on much less observations, the effects of information density might have been amplified.\footnote{We also ran an analysis including only verbs used in \citet{jaeger2010redundancy}, and its results are qualitatively similar to our full analysis, suggesting that it is indeed the size of the dataset (i.e., more observations per verb) that leads to different results.}

Beyond UID, we also find support for other accounts of \textit{that}-mentioning. First, lower-frequency matrix verbs and CC onsets are associated with more \textit{that}-mentioning, consistent with availability-based accounts. We also find syntactic priming: speakers are more likely to include \textit{that} if the previous CC did. Evidence for dependency locality is mixed—longer CC remainders increase \textit{that}-mentioning, but greater distance between the matrix verb and CC onset, as well as longer CC subjects, showed the opposite pattern. This may be due to parsing errors or shifting usage patterns. Effects of speaker commitment are also observed, with higher levels of speaker commitment leading to less overt \textit{that}. Finally, we observe similarity avoidance (reduced \textit{that}-use in potential \textit{that}-\textit{that} sequences) and disfluency effects (filled words and repetitions increased \textit{that}-mentioning).

Our findings also shed light on the structural sensitivity of GPT-2, particularly its contextual word embeddings. Embeddings of the matrix verb—derived solely from pre-CC context are predictive of upcoming syntactic structure, suggesting that GPT-2 embeddings captures fine-grained structural cues. This may explain the success of GPT-2 and other autoregressive neural language models in downstream tasks that require syntactic knowledge. This approach offers a promising avenue for future work to leverage contextual embeddings for modeling syntactic prediction more broadly.

\section{Conclusion}
This study provides robust support for UID at the syntactic level in naturalistic conversations. Information density estimated from contextual word embeddings significantly predicted \textit{that}-mentioning, even after controlling for verb-specific preferences. Additionally, we showed that verb-specific preferences also played an important role, and that information density measures derived from verbs' subcategorization probabilities might have been conflated with verb-specific preferences. These findings highlight limitations of conventional linguistic features in modeling predictive processes, and suggest that high-dimensional linguistic representations such as contextual word embeddings offer a more effective and flexible alternative. Our results also demonstrate that \textit{that}-reduction is shaped by multiple interacting pressures—including information density, availability, speaker commitment, syntactic priming, and form avoidance.

Lastly, our work underscores the value of combining large naturalistic corpora with machine learning and NLP techniques for studying psycholinguistics. The use of the CANDOR corpus allowed us to examine \textit{that}-mentioning in spontaneous, naturalistic speech across a diverse linguistic samples. By leveraging machine learning and contextual word embeddings from neural language models, we developed more nuanced predictors of structural choices. This approach not only improves predictive accuracy but also opens new avenues for modeling linguistic behavior at scale. 

\section*{Limitations}

There are several limitations to the present study. First, the conversational transcripts were automatically generated, and dependency structures were derived using automatic parsers. As a result, the data may contain transcription and parsing errors. Second, we relied on GPT-2 to estimate online spoken language predictions, although GPT-2 is primarily trained on written text. This may limit its ability to fully capture characteristics of spontaneous spoken language. Moreover, our analysis was based on a single language model architecture. Future work should explore alternative models, including those trained on conversational data or designed for speech-oriented tasks, to assess the generalizability of our findings. Lastly, although our analysis found that no linguistic features significantly improved the structural predictability of CCs, it is possible that we did not exhaust the full range of relevant linguistic predictors. Future research could investigate additional features that may contribute to the reduction of \textit{that} in CCs.

\section*{Ethical Considerations}

We employed AI-based tools (Claude and ChatGPT) for writing and coding assistance. These tools were used in compliance with the ACL Policy on the Use of AI Writing Assistance.

%\bibliographystyle{acl_natbib}

%\bibliographystyle{acl_natbib}
\bibliography{anthology,custom}

\appendix

\section{Matrix Verb Statistics}
\label{sec:verbs}

This appendix provides the full distribution of complement clause subcategorization probabilities across verbs in our dataset in Table~\ref{tab:appendix_subcat_table_part1} and Table~\ref{tab:appendix_subcat_table_part2}.

\begin{table*}[t]
\centering
\begin{tabular}{lrrr}
\hline
\textbf{Verb Lemma} & \textbf{Total Occurrences} & \textbf{CC Occurrences} & \textbf{Subcat Probability (\%)} \\
\hline
know & 119,678 & 28,664 & 23.95 \\
think & 46,610 & 35,080 & 75.26 \\
mean & 30,281 & 1,916 & 6.33 \\
say & 24,805 & 13,612 & 54.88 \\
like & 23,578 & 3,381 & 14.34 \\
see & 20,578 & 7,111 & 34.56 \\
take & 15,314 & 994 & 6.49 \\
feel & 11,274 & 2,298 & 20.38 \\
guess & 9,744 & 6,101 & 62.61 \\
hear & 9,166 & 2,408 & 26.27 \\
tell & 7,264 & 3,345 & 46.05 \\
find & 6,579 & 1,948 & 29.61 \\
love & 6,290 & 762 & 12.11 \\
thank & 5,521 & 289 & 5.23 \\
remember & 4,626 & 2,191 & 47.36 \\
read & 3,649 & 346 & 9.48 \\
show & 3,170 & 650 & 20.50 \\
understand & 2,984 & 1,092 & 36.60 \\
suppose & 2,911 & 326 & 11.20 \\
hope & 2,488 & 1,869 & 75.12 \\
teach & 2,380 & 238 & 10.00 \\
figure & 2,327 & 912 & 39.19 \\
believe & 1,970 & 947 & 48.07 \\
imagine & 1,891 & 874 & 46.22 \\
\hline
\end{tabular}
\caption{Verb-level complement clause frequencies and subcategorization probabilities (part 1).}
\label{tab:appendix_subcat_table_part1}
\end{table*}

\begin{table*}[t]
\centering
\begin{tabular}{lrrr}
\hline
\textbf{Verb Lemma} & \textbf{Total Occurrences} & \textbf{CC Occurrences} & \textbf{Subcat Probability (\%)} \\
\hline
check & 1,754 & 114 & 6.50 \\
care & 1,693 & 263 & 15.53 \\
decide & 1,428 & 579 & 40.55 \\
realize & 1,395 & 974 & 69.82 \\
agree & 1,324 & 172 & 12.99 \\
hold & 1,313 & 107 & 8.15 \\
wish & 1,291 & 1,028 & 79.63 \\
worry & 1,028 & 90 & 8.75 \\
expect & 980 & 349 & 35.61 \\
consider & 840 & 264 & 31.43 \\
mind & 733 & 208 & 28.38 \\
notice & 721 & 324 & 44.94 \\
mention & 645 & 190 & 29.46 \\
answer & 561 & 26 & 4.63 \\
explain & 561 & 106 & 18.89 \\
bet & 480 & 272 & 56.67 \\
accept & 465 & 49 & 10.54 \\
complain & 423 & 53 & 12.53 \\
stress & 234 & 23 & 9.83 \\
admit & 209 & 98 & 46.89 \\
respond & 176 & 11 & 6.25 \\
joke & 156 & 32 & 20.51 \\
promise & 146 & 58 & 39.73 \\
judge & 119 & 19 & 15.97 \\
claim & 110 & 47 & 42.73 \\
suggest & 108 & 50 & 46.30 \\
\hline
\end{tabular}
\caption{Verb-level complement clause frequencies and subcategorization probabilities (part 2).}
\label{tab:appendix_subcat_table_part2}
\end{table*}

\section{Descriptive Statistics for Predictors for Modeling \textit{that}-Mentioning}
\label{sec:variables}

This appendix provides the full distribution of complement clause subcategorization probabilities across verbs in our dataset in Table~\ref{tab:predictors}.

\begin{table*}[t]
\centering
\begin{tabular}{p{4cm}p{3cm}p{6.5cm}}
\hline
\textbf{Predictor} & \textbf{Type} & \textbf{Values / Distribution} \\
\hline
CC Subject Frequency & Continuous & -- \\
CC Subject Form & Categorical (4 levels) & I (29.62\%), You (13.15\%), other pronouns (41.80\%), other NPs (15.42\%) \\
Matrix Verb Frequency & Continuous & -- \\
Co-referentiality & Binary & yes (32.72\%), no (67.28\%) \\
Previous \textit{that} & Binary & present (11.46\%), absent (88.54\%) \\
Matrix Verb-CC Distance & Binary & local (84.73\%), non-local (16.27\%) \\
CC Subject Length & Continuous & -- \\
CC Reminder Length & Continuous & -- \\
Matrix Subject Form & Categorical (4 levels) & I (72.70\%), You (10.37\%), other pronouns (13.49\%), other NPs (3.44\%) \\
Position & Continuous & -- \\
\textit{that}-Doubling & Binary & present (3.03\%), absent (96.97\%) \\
Filled Word & Binary & present (10.32\%), absent (89.68\%) \\
Repetition & Binary & present (4.12\%), absent (95.88\%) \\
\hline
\end{tabular}
\caption{Overview of predictors included in the statistical model, along with their types and distribution where applicable.}
\label{tab:predictors}
\end{table*}

\section{By-verb Plot for Effects of Verb-based Information Density on \textit{that}-mentioning}

\begin{figure*}[ht]
  \centering
  \includegraphics[width=\textwidth]{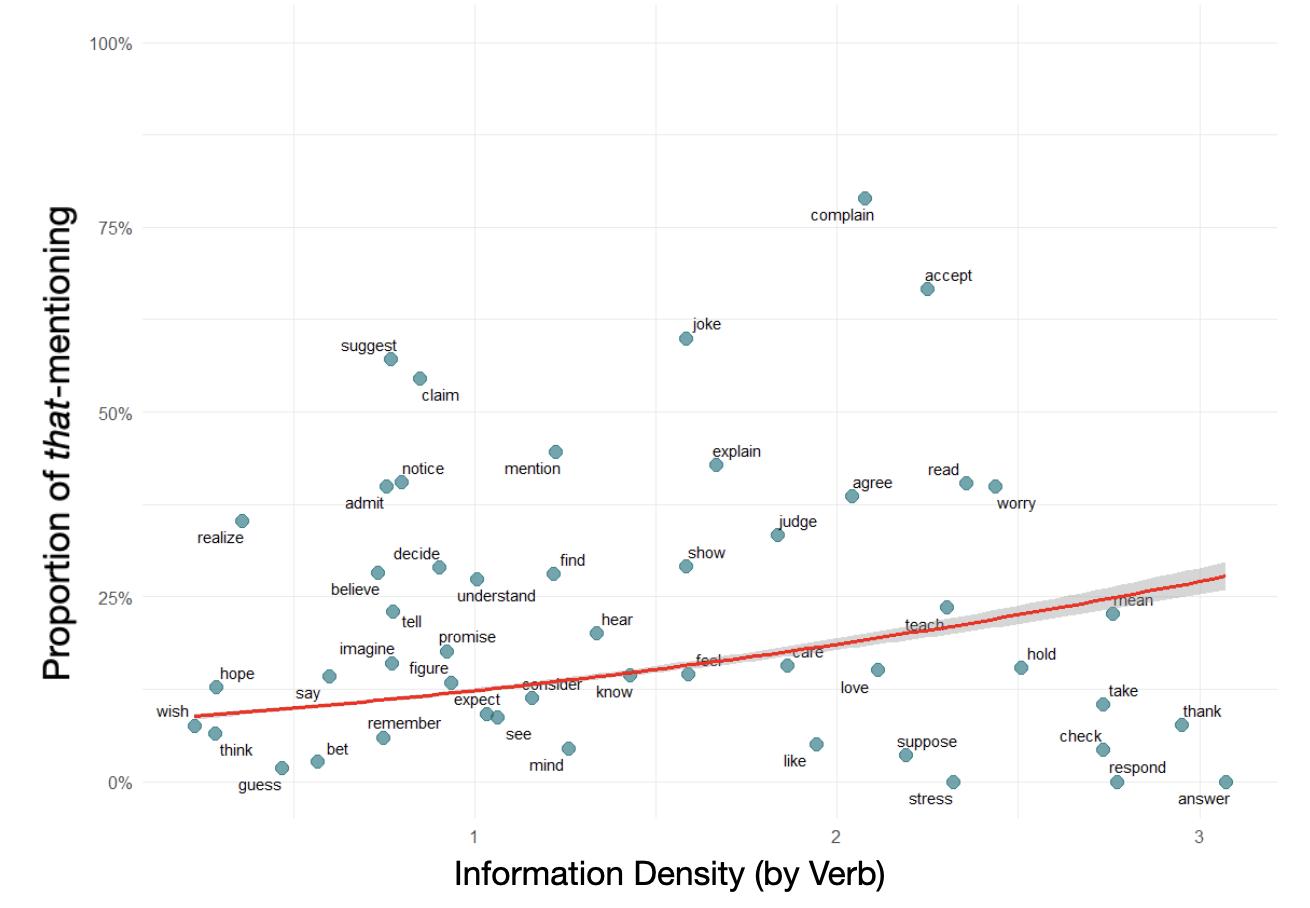}
  \caption{Effects of information density (by verb Subcategorization Probability) on \textit{that}-mentioning. The red line is a logistic regression fit estimated as a binomial GLMM. Each dot represents a verb.}
  \label{fig:withverbs}
\end{figure*}

In Figure~\ref{fig:withverbs} we plot the relationship between verb-based information density and \textit{that}-mentioning, including the identities of verbs. 

\label{sec:byverb}
\clearpage

\end{document}